\documentclass{bmvc2k}

\usepackage{xspace}
\usepackage{paralist}
\usepackage{booktabs}
\usepackage{wrapfig}

\DeclareGraphicsExtensions{.pdf,.jpg,.png}
\graphicspath{{figs/}}

\newcommand{\secref}[1]{Section~\ref{sec:#1}}
\newcommand{\figref}[1]{Figure~\ref{fig:#1}}
\newcommand{\tblref}[1]{Table~\ref{tbl:#1}}
\newcommand*{\eg}{e.g.\@\xspace}
\newcommand*{\ie}{i.e.\@\xspace}

\def\eg{\emph{e.g}\bmvaOneDot}

\title{Horizon Lines in the Wild}

\addauthor{Scott Workman}{scott@cs.uky.edu}{1}
\addauthor{Menghua Zhai}{ted@cs.uky.edu}{1}
\addauthor{Nathan Jacobs}{jacobs@cs.uky.edu}{1}

\addinstitution{
 Department of Computer Science\\
 University of Kentucky\\
 Lexington, KY, USA 
}

\runninghead{Workman, Zhai, Jacobs}{Horizon Lines in the Wild}

\begin{document}

\maketitle

\begin{abstract}

The horizon line is an important contextual attribute for a wide
variety of image understanding tasks. As such, many methods have been
proposed to estimate its location from a single image. These methods
typically require the image to contain specific cues, such as
vanishing points, coplanar circles, and regular textures, thus limiting
their real-world applicability.  We introduce a large, realistic
evaluation dataset, {\em Horizon Lines in the Wild} (HLW), containing
natural images with labeled horizon lines.  Using this dataset, we
investigate the application of convolutional neural networks for
directly estimating the horizon line, without requiring any explicit
geometric constraints or other special cues. An extensive evaluation
shows that using our CNNs, either in isolation or in conjunction with
a previous geometric approach, we achieve state-of-the-art results on
the challenging HLW dataset and two existing benchmark datasets.

\end{abstract}

\section{Introduction}

Single image horizon line estimation is one of the most fundamental
geometric problems in computer vision. Knowledge of the horizon line
enables a wide variety of applications, including: image
metrology~\cite{criminisi2000single}, geometrically biased pedestrian
and vehicle detection~\cite{hoiem2008putting}, and perspective
correction in consumer photographs~\cite{lee2012automatic}. 
Despite this demonstrated importance, progress on this task
has stagnated and nearly all recent methods that focus on
this problem make assumptions about the presence of particular
geometric objects in the scene, such as vanishing
points~\cite{tardif2009non,xu2013minimum,lezama2014finding},
repeated textures~\cite{criminisi2000texture}, and coplanar
circles~\cite{chen2004camera}. Existing
benchmark datasets for single image horizon line
estimation~\cite{barinova2010geometric,denis2008efficient} were
created to evaluate methods that use the orthogonal vanishing point
cue, contributing to this
stagnation. 

We introduce a new benchmark dataset, {\em Horizon Lines in the Wild} (HLW),
containing real-world images with labeled horizon lines. Our dataset
is significantly larger and more diverse than existing benchmark
datasets for horizon line detection.  Instead of focusing on a
particular geometric cue, we take a learning-based approach and
propose to use a deep convolutional neural network (CNN) to directly
estimate the horizon line. The resulting network implicitly combines
both geometric and semantic cues, makes no explicit assumptions about
the contents of the underlying scene, and is several orders of
magnitude faster than current state-of-the-art methods which focus on
vanishing points. 


Recent work using learning-based methods for horizon line estimation
has been limited, with three notable exceptions. Fefilatyev et
al.~\cite{fefilatyev2006horizon} proposed to segment the sky and then
detect the horizon line in the resulting binary mask. This approach is
limited to when the horizon line is visible, such as from a boat on
the ocean on a clear day. Ahmad et al.~\cite{ahmad2013machine}
proposed a segmentation approach to estimate the location of the
skyline, a closely related, but distinct, problem. Zhai et
al.~\cite{zhai2016context} use a CNN as a prior over likely
horizon line locations, but they focus on the vanishing point cue. We
propose to use a CNN to directly estimate the horizon line location.
However, we show that by using our CNN as context for their method,
replacing the one they proposed, significantly improves
performance for vanishing point based horizon line
estimation.  Extensive experiments demonstrate that our CNN-based
approach is fast, requiring only milliseconds per image, and accurate,
achieving state-of-the-art performance on two popular datasets
designed to showcase purely geometric methods, and the challenging HLW
dataset. 

Our main contributions are: 1) a novel approach for using structure
from motion to automatically label images with a horizon line, 2) a
large evaluation dataset of images with labeled horizon lines, 3) a
CNN-based approach for directly estimating the horizon line in a
single image, and 4) an extensive evaluation of a variety of CNN
design choices.

\subsection{Horizon Line: Geometric Definition}

The image location of the horizon line is defined as the projection of
the line at infinity for any plane which is orthogonal to the local
gravity vector. The gravity vector often coincides with the local
ground plane surface normal, but not always.  This is distinct from
the problem of detecting the skyline, which is the set of points
where the sky and the ground meet. 
    
A camera is defined by its extrinsic and intrinsic parameters.  A
point in the world, $X_i$, is related to a pixel, $p_{ci}$, in
a camera, $c$, as follows: 
\begin{equation}
  [u_{ci},v_{ci},1]^\mathsf{T} = p_{ci} \propto K_c(R_c X_i + t_c),
\end{equation}
where $R_c$ is the camera orientation, $t_c$ is the camera
translation, and $K_c$ is the intrinsic calibration.  For our camera
coordinates we assume that the positive $x$-direction is to the right,
the positive $y$-direction is up, and the viewing direction is down
the negative $z$-axis. Using this parameterization, the world viewing
direction of our camera is $R_c^\mathsf{T}[0,0,-1]^\mathsf{T}$.
Assuming that the world vector $[0,1,0]^\mathsf{T}$ points in the
zenith direction, the horizon line in our image is defined as the set
of pixels, $p$, where 
\begin{equation}
  p^\mathsf{T}K_c^{-T}R_c[0,1,0]^\mathsf{T} = 0.
\end{equation}
If the intrinsic calibration, $K_c$, of the camera is known, then the
horizon line provides a sufficient set of constraints to estimate the
camera tilt and roll in world coordinates.

\section{A New Dataset for Horizon Line Detection}

We introduce {\em Horizon Lines in the Wild} (HLW), a large dataset of
real-world images with labeled horizon lines, captured in a diverse
set of environments. 
The dataset is available for download at
our project website~\cite{hlwsite}.
We begin by characterizing limitations in existing
datasets for evaluating horizon line detection methods and then
describe our approach for leveraging structure from motion to
automatically label images with horizon lines.

\subsection{Limitations of Existing Datasets}

\begin{figure}

  \centering
 
  \subfigure[ECD]{
    \includegraphics[width=.315\linewidth]{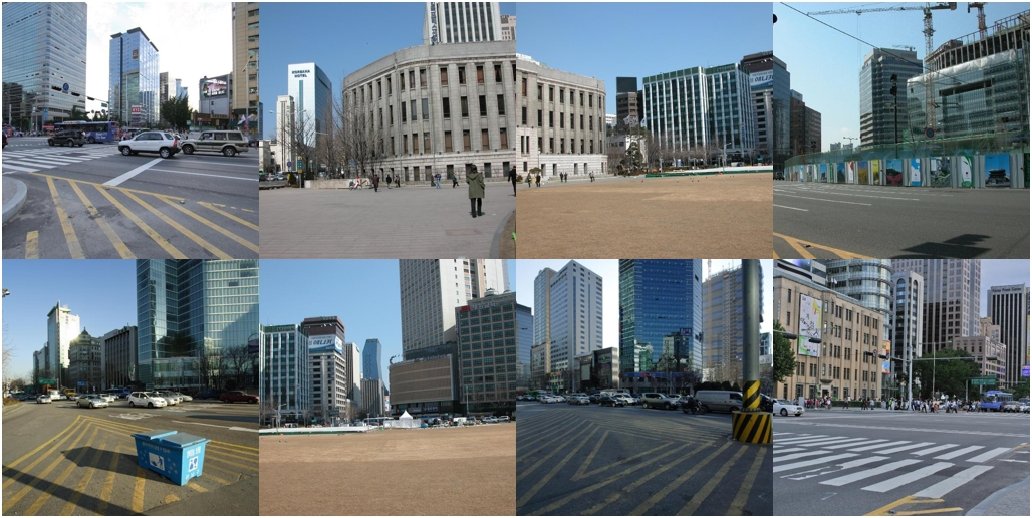}
  }
  \subfigure[HLW]{
    \includegraphics[width=.315\linewidth]{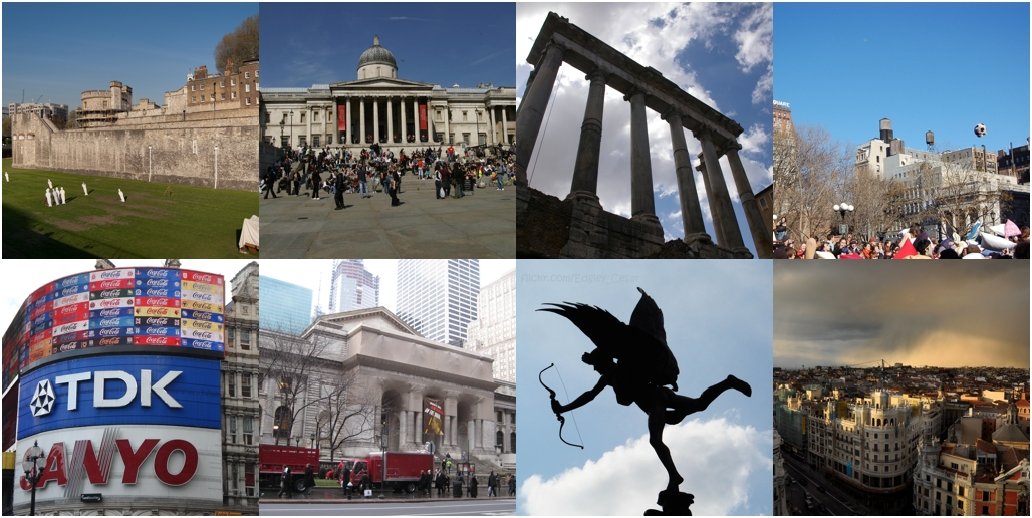}
  }
  \subfigure[HLW + Street-Side]{
    \includegraphics[width=.315\linewidth]{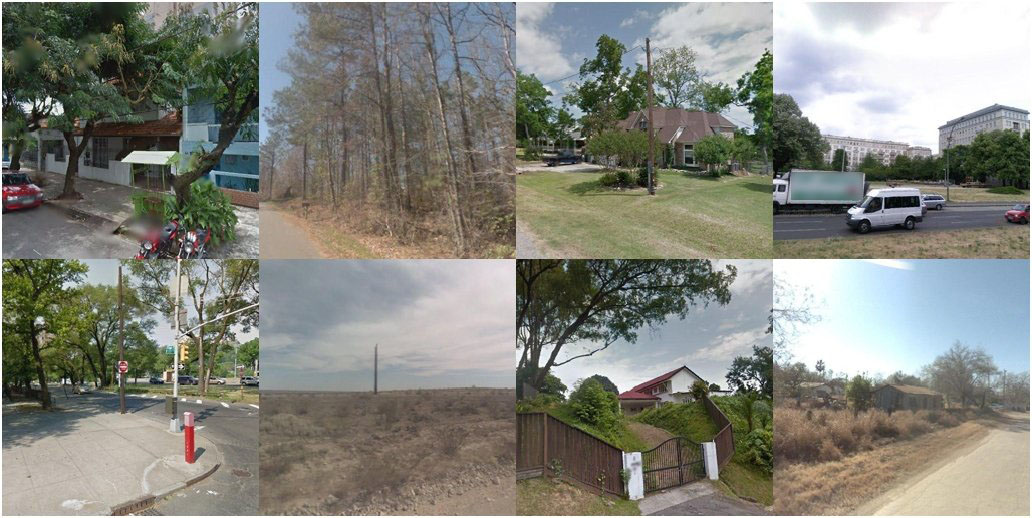}
  }

  \caption{Montages highlighting the diversity of perspectives and scenes in HLW.} 

  \label{fig:dataset}

\end{figure}

There are two main datasets that have been used in recent work on
estimating horizon lines: the Eurasian Cities
Dataset~\cite{barinova2010geometric} (ECD) and the older York Urban
Dataset~\cite{denis2008efficient} (YUD). We argue that these datasets
have outlived their usefulness. They are too small and do
not reflect the diversity of environments in which real-world
horizon line detection methods must work.

ECD is the predominant benchmark dataset used for evaluating automatic
vanishing point detection algorithms. It consists of 103 outdoor
images captured in large urban areas, many of which do not satisfy the
Manhattan world assumption~\cite{coughlan1999manhattan}, \ie, that
most lines correspond to one of three mutually orthogonal directions,
one of which is up. Of these images, the first 25 are used for model
fitting, with the remainder used for testing. Of the 78 testing
images, a majority are considered quite easy. Due to a combination of
the few number of testing images and the small number of
challenging images, the difference in performance between various
methods often depends on a single image.
The older YUD dataset is similarly small (102 images, first 25 for
model fitting) and is seen as too easy because the images are captured in a
confined area with a single camera, there are relatively fewer outlier
line segments, the scenes satisfy the Manhattan world assumption, and
there is no camera roll.  

To obtain ground truth horizon lines for ECD and YUD, a manual process
akin to the following was used: identify families of parallel line
segments, estimate a vanishing point for each, and compute the horizon
line from the horizontal vanishing points using a least squares fit.
This process is slow, error prone, and severely limits the diversity
of scenes. As Lezama et al.~\cite{lezama2014finding} note, there is
even a duplicated testing image in ECD, with each instance having a
different ground truth horizon line. 

It is our belief that the limitations of these datasets have caused
useful progress in this research area to stagnate. Recent 
state-of-the-art methods are quite slow, which is reasonable when you
have a small testing dataset.  For example, we find that the
approach of Lezama et al.~\cite{lezama2014finding} requires
approximately 30 seconds per image on YUD and 1 minute per image on
ECD (results obtained using code made available by the authors). These
methods have also focused on a particular processing pipeline: detect
line segments, find vanishing points, then globally optimize to find a
consistent scene interpretation. The reliance on vanishing points
limits these methods to regions with many man-made structures. There
is clearly a need for a larger and more diverse dataset for evaluating
horizon line estimation methods.

\subsection{Leveraging Structure from Motion}
\label{sec:sfm}

\begin{figure}
 
  \centering
 
  \subfigure[]{
    \includegraphics[height=.32\linewidth,trim={5.3cm 5.4cm 4cm 4.5cm},clip]{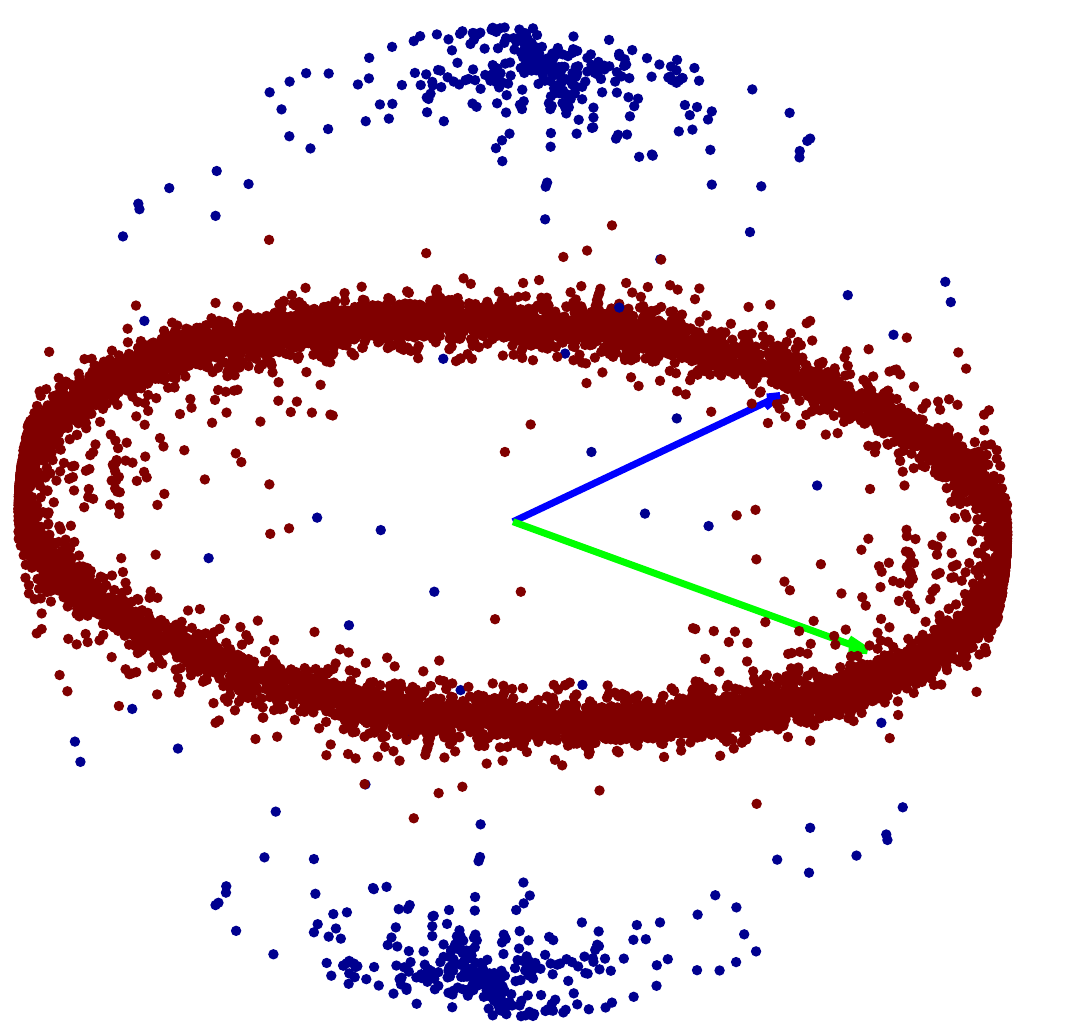}
  }
  \subfigure[]{
    \includegraphics[height=.32\linewidth]{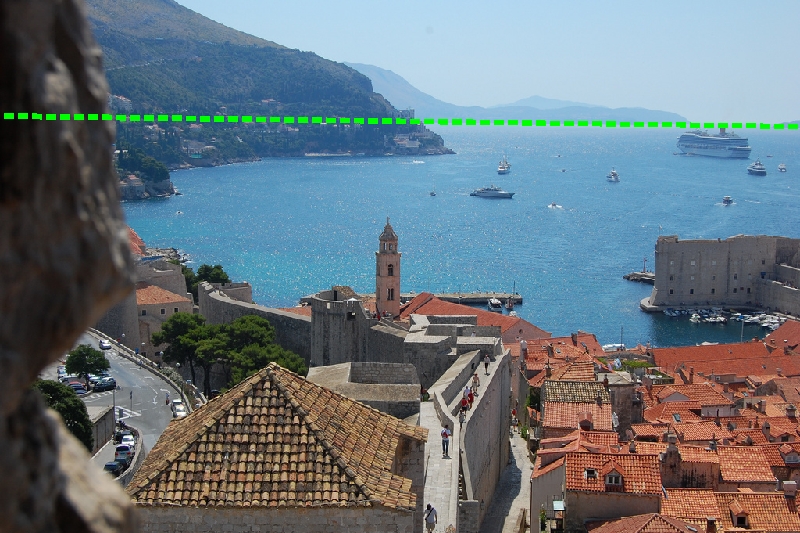}
  }

  \caption{Using a SfM model to estimate the horizon line. (a) Each
  point represents the left/right direction of an image in world
  coordinates (blue = outlier). Two vectors represent the estimated
  horizon plane. (b) The horizon line projected into one image from
  the model.}
  
  \label{fig:sfm}
\end{figure}

We introduce a novel technique for automatically
labeling images with horizon lines using structure from motion (SfM),
which we then employ to generate a large evaluation dataset. Kendall
et al.~\cite{kendall2015convolutional} used a similar strategy to
generate a dataset to evaluate a CNN-based method for camera
relocalization. Their work focused on learning a scene-specific CNN,
whereas our goal is a scene-agnostic CNN that does not require
scene-specific training data.

The output of SfM is the extrinsic and intrinsic camera parameters for
a subset of the input images. Typically these images are downloaded
from photo-sharing websites, such as Flickr, around major landmarks.
The extrinsic coordinates output by SfM algorithms typically have an
unknown global orientation and translation. Since our focus is the
horizon line, we just need to estimate the global up direction (the
yaw of the reconstruction is irrelevant to our needs).  A commonly
used 
approach to estimate the global orientation is to average the image
`up' directions in world coordinates.  The implicit assumption of this
approach is that the expected tilt and roll of a camera is zero.
While this works well in many cases, it fails in scenes with a single
dominant landmark that is viewed from one direction (\eg, Notre Dame
in Paris). 
In practice, we found that we get more reliable world zenith direction
estimates if we instead only assume that the expected roll of a camera
is zero. For a given set of images, we solve for the world direction of
the points at infinity in the left, $[-1,0,0]$, and right, $[1,0,0]$,
directions. Given a set of these points, we use singular value
decomposition to estimate a basis for the horizon plane
(\figref{sfm}), ignoring images that are rotated by 90 degrees (using
reconstruction error).

Starting from 185 high-quality SfM models in the
1DSfM~\cite{wilson2014robust}, Landmarks~\cite{li2012worldwide}, and
YFCC100M~\cite{heinly2015reconstructing} datasets, we filtered out
anomalous images, fit and manually validated a global horizon line for
each model, and then projected the horizon line back into each image. The
resulting dataset, HLW, contains $100\,553$ images. From
each 1DSfM model, we hold out 100 images at random, including holding
out two models completely, resulting in $2\,018$ images to be used for
evaluation. We hold out 525 training images for validation (approximately 3
from each model). 

\subsection{Augmenting using Street-Side Imagery}

The SfM models are mostly of tourist landmarks, which are usually in
urban areas. Images of more natural areas, such as Mount Rushmore,
Stonehenge, and the Grand Canyon, are included.  However, the dataset
contains few, if any, images of many scene types, including: forests,
crop fields, industrial parks, and residential streets. To reduce this
bias, we augment our training dataset with rectilinear cutouts
extracted from equirectangular street-side imagery panoramas (via
Google Street View).  

We first use the SfM models to learn a
plausible distribution of camera focal length (equivalently field of
view), tilt, and roll.  We model focal length using a normal
distribution. We find that the camera roll is well modeled by
Student's t-distribution ($v=2.43$). For camera tilt, we use a kernel
density estimate (Epanechnikov kernel, $\sigma=.003$).  Camera yaw is
sampled uniformly at random. Starting from $50\,000$ panoramas,
sampled from the continental US and 93 metropolitan areas around the
world, we generate $500\,000$ training images by randomly sampling
square cutouts based on the learned distributions. 

\subsection{Comparisons with Existing Datasets}
\label{sec:comparison}

\begin{figure}

  \centering
 
  \subfigure[ECD]{
    \includegraphics[width=.231\linewidth]{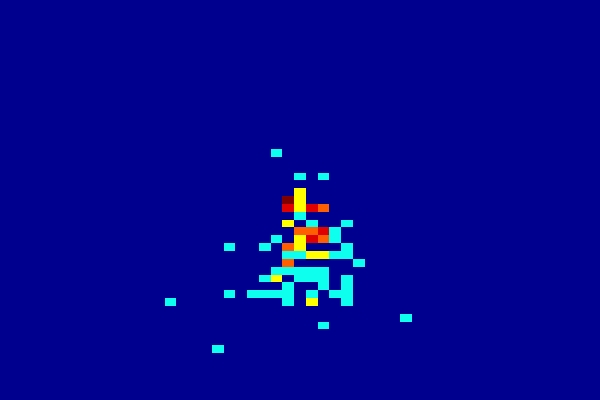}
  }
  \subfigure[YUD]{
    \includegraphics[width=.231\linewidth]{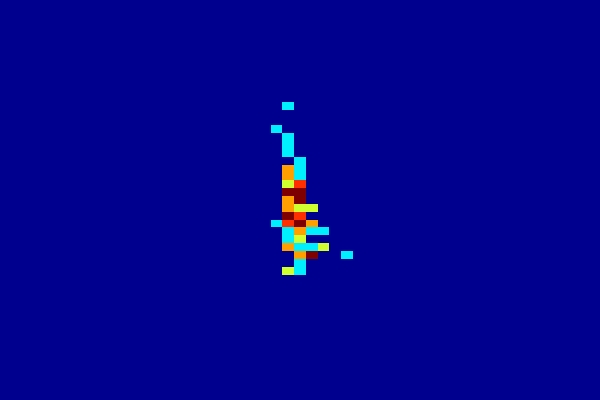}
  }
  \subfigure[HLW]{
    \includegraphics[width=.231\linewidth]{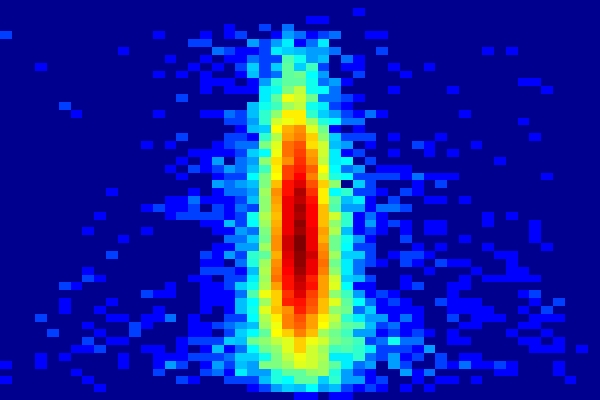}
  }
  \subfigure[HLW + Street-Side]{
    \includegraphics[width=.231\linewidth]{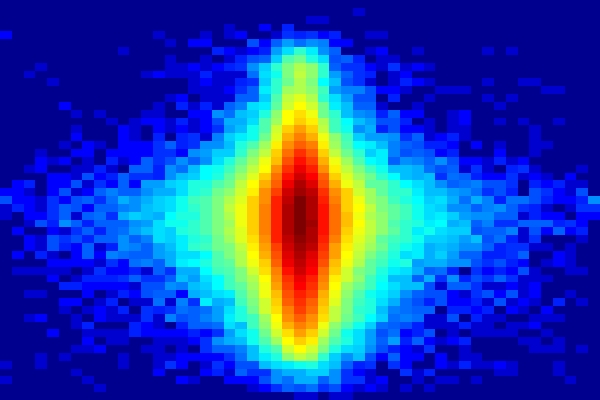}
  }

  \caption{Distribution of horizon lines for images in HLW versus
  other benchmark datasets (red = higher likelihood). The $x$-axis is
slope and the $y$-axis is vertical offset.}

  \label{fig:horizon_dists}

\end{figure}

\begin{wrapfigure}{R}{.54\linewidth}
  
  \centering
  
  \includegraphics[width=1\linewidth]{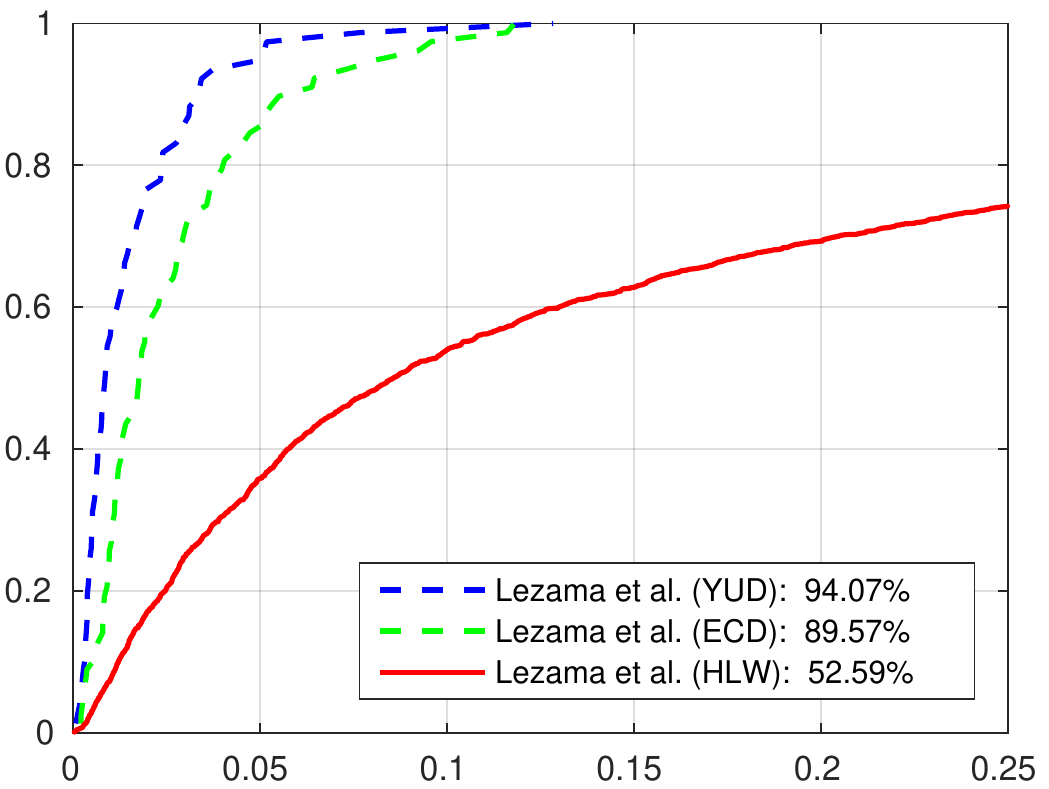}

  \caption{Evaluating the recent state-of-the-art method by
    Lezama et al.~\cite{lezama2014finding} on HLW. The fraction of
    images ($y$-axis) with a horizon error less than a threshold
    ($x$-axis). The AUC is shown in the legend.}
 
  \label{fig:hlw_results}
\end{wrapfigure}

A montage of sample images from HLW and ECD are shown in \figref{dataset}. Even when considering this
small set of images, there is clearly much greater diversity of scene
types in HLW (\eg, zoomed in view of a statue, elevated view of a
city). The scenes in ECD consist primarily of urban images with large
buildings in the background. HLW also has a wider and much more
densely sampled distribution of horizon line locations than ECD or
YUD. We represent the horizon line as $\rho = x \cos \theta + y \sin
\theta$, where $\rho$ is the perpendicular distance from the origin to
the horizon line and $\theta$ is the angle the horizon line makes with the horizontal
axis. \figref{horizon_dists} shows the joint distribution over
$\theta$ ($x$-axis) and $\rho$ ($y$-axis) for each dataset.

We evaluated the recent state-of-the-art method by Lezama et
al.~\cite{lezama2014finding} on HLW. The standard error metric used
for horizon line detection is the maximum distance from the detection
to the ground truth in image space, normalized by the height of the
image, which we refer to as horizon detection error. This is often
reported for a set of images as the area under the curve of the
cumulative histogram of errors (AUC).  Barinova et
al.~\cite{barinova2010geometric} motivate the use of horizon detection
error as the standard accuracy measure for automatic vanishing point
detection algorithms. \figref{hlw_results} visualizes the result. The
large relative performance difference compared to other benchmarks
highlights the challenging nature of the HLW dataset.

\section{Direct Horizon Line Estimation}
\label{sec:methods}

We propose to use convolutional neural networks (CNNs) to estimate the
location of the horizon line from raw pixel intensities. This approach
is fast and does not require extensive manual tuning of parameters.
Importantly, the computational cost only depends on the size of the
image, not the content of the scene, such as the number of line
segments. Our work explores design and implementation choices which
have a significant impact on the accuracy of the resulting model,
including: target label space, weight initialization, and objective
function.


For all experiments we use the GoogleNet
architecture~\cite{szegedy2015going} because it achieves similar
accuracy to other architectures we tested, but with many fewer
parameters. Our CNNs expect the input images to have a fixed size and
a square aspect ratio.  For non-square images, we extract a maximal
square center crop and, optionally, a dense
grid if using an aggregation strategy (\secref{merging}). We
experimented with reshaping the image to be square, but the resulting
networks were far less accurate. This result is in line with previous
work~\cite{workman2015deepfocal} showing that maintaining aspect
ratio is important when estimating camera focal length, which is a
closely related geometric task.

We consider two parameterizations of the
horizon line: 1) slope/offset, $(\theta,\rho)$, where $\rho$ is the
perpendicular distance from the origin to the horizon line and
$\theta$ is the angle the horizon line makes with the $x$-axis of the image and 2)
left/right, $(l,r)$, where $l$ is the vertical offset at which the
horizon line intersects the left side of the image, $r$ is similarly
defined. We represent $\rho$, $l$, and $r$ in units of image heights.
The remainder of this section describes two CNN variants for
predicting the horizon line location.

\subsection{Classification Approach}
\label{sec:classification}

As most existing work has applied CNNs for classification tasks, we
initially frame horizon line estimation as a classification problem.
The primary benefit of such a formulation is that the output of a CNN
trained for a one-of-many classification task is a probability
distribution over the categories; in our case a distribution over
possible horizon lines in the image. For each parameter we generate
$N=100$ bins by linearly interpolating the cumulative distribution
function of that parameter over the training data. Additionally for
slope, $\theta$, we force the bin edges to be symmetric. 

Our process for adapting the GoogleNet architecture is as follows: 1)
duplicate each softmax classifier (a fully connected layer followed by
a multinomial logistic loss, where real-valued predictions are first
passed through a softmax function to get a probability distribution
over classes) to occur once for each parameter and then 2) modify the
fully connected layer for each softmax classifier to output a
$N$-dimensional vector corresponding to the $N$ bins.


\subsection{Regression Approach}
\label{sec:regression}

Regression using deep CNNs is widely seen as more challenging than
classification due to difficulties in controlling the optimization
process and handling outliers.
Despite this, recent work has proposed to use deep CNNs for regression
tasks, including: pose estimation~\cite{tompson2015efficient}, camera
relocalization~\cite{kendall2015convolutional}, and depth
estimation~\cite{eigen2014depth}. As discussed by Belagiannis
et al.~\cite{belagiannis2015robust}, optimization is typically
performed using the $L_2$ loss, but outliers reduce the generalization
ability of the network and increase the convergence time.
Girshick~\cite{girshick2015fast} note that if the regression targets
are unbounded, training with the $L_2$ loss can require careful
parameter tuning to prevent exploding gradients. 


For our regression networks we minimize the Huber
loss~\cite{huber1964robust}, a robust loss function that is less
sensitive to outliers:
\begin{equation}
  L(x) = \begin{cases} 
    \frac{1}{2}x^2 & \text{for } |x| \leq \delta, \\
    \delta(|x| - \frac{1}{2}\delta) & \text{otherwise}.
  \end{cases}
\end{equation}
For this work, we set $\delta = 1$. To adapt the GoogleNet
architecture for regressing the horizon line, we replace each softmax
classifier with a regressor (once for each parameter) and modify the
corresponding fully connected layer to output a scalar.

Our results show that optimization using the Huber loss results in
more accurate predictions than using the $L_2$ loss. However, using
only a regression objective did not perform as well as a
classification objective. To address this, we investigated two initialization
strategies: 1) initializing from the weights of a previously trained
classification network, and 2) jointly optimizing a classification and
regression network, with shared weights, where the softmax classifiers
act as a form of regularization. We find that using both strategies,
we can significantly improve performance and reduce convergence time,
even when using the $L_2$ loss.

\subsection{Aggregating Estimates Across Subwindows}
\label{sec:merging}

When applied to classification problems, the standard procedure for
processing an image through a CNN is to extract multiple subwindows,
feed each through the network separately, and average the predictions.
This strategy is applicable to the problem of object recognition,
where the target label is shared across subwindows. For horizon line
estimation, each subwindow has a unique target label (as the
horizon line position changes). Therefore this strategy is
insufficient. 

We propose two strategies for aggregating estimates: 1) projecting the
horizon line from the subwindow to the full-size image and averaging
in image space (weighted by the confidence in each estimate), and 2)
optimizing for the horizon line in the full image that is maximally
likely in all subwindows. For the latter, we assume that each
subwindow is independent and minimize the negative log-likelihood, 
\begin{equation}
  E = -\frac{1}{N}\sum_{i=1}^{N}log(W(I_i;\Theta)), 
\end{equation}
where $W$ is a function that maps the global horizon line, $\Theta$,
into the coordinate frame for subwindow $I_i$, and extracts the
probability.  Our results show that both strategies improve accuracy
relative to using only a center crop, but the averaging strategy is
faster. 


\section{Experiments}
\label{sec:bakeoff}

We conducted an extensive evaluation of our proposed techniques, which
use convolutional neural networks for horizon line estimation, on the
HLW, YUD, and ECD datasets. By using our networks, either in isolation
or in conjunction with a previous method, we achieve state-of-the-art
results on all datasets.


\subsection{Implementation Details}

We implemented the proposed networks using the Caffe~\cite{jia2014caffe}
deep learning toolbox. Sample code, including models and solver
settings, is available on the project website~\cite{hlwsite}.
We trained each network using stochastic gradient descent with
a step learning rate policy, a mini-batch size of 40, for $125\,000$
iterations (approximately $35$ epochs). We set the base learning rates
to $10^{-3}$ and $10^{-5}$ for classification and regression,
respectively, decreasing by an order of magnitude every $25\,000$
iterations (when training from scratch, we use the GoogleNet {\em
quick solver} ~\cite{jia2014caffe}). We kept a snapshot every $1\,000$
iterations, selecting the snapshot that minimizes horizon error on the
HLW validation set. The input image size for all of our networks is
$224 \times 224$. 

We combined the HLW and street-side imagery to form a training set. For
the HLW imagery, we performed data augmentation by randomly mirroring the image
horizontally with 50\% probability and sampling a square crop (minimum
side length $85\%$ of the smallest image dimension). We extracted ten
crops from each image, adjusting the horizon line for each. Since the
street-side imagery was already square with randomly sampled camera
orientations, we just scaled to the input size of the network. 

\subsection{Quantitative Evaluation}

\begin{table}
  \centering
  \caption{Evaluation of our networks on HLW and ECD.}
  \begin{tabular}{@{}llcccccc@{}}
    \toprule
    & Loss & \multicolumn{2}{c}{HLW ({\em held})} & \multicolumn{2}{c}{HLW ({\em all})} & \multicolumn{2}{c}{ECD}\\
    &      & $(\theta, \rho)$ & $(l, r)$    & $(\theta, \rho)$ & $(l, r)$    & $(\theta, \rho)$ & $(l, r)$ \\
    \hline
    \multicolumn{8}{@{}l@{}}{\textbf{Classification}} \\ 
    ~~ImageNet & Softmax   & 64.49\% & 62.10\%   & 69.02\% & 67.08\%   & 82.28\% & 82.99\% \\ 
    ~~Places   & Softmax   & 65.73\% & 59.54\%   & 69.97\% & 67.38\%   & 83.96\% & 80.45\% \\ 
    ~~PoseNet  & Softmax   & 60.49\% & 61.35\%   & 61.65\% & 63.56\%   & 78.36\% & 77.77\% \\ 
    ~~Salient  & Softmax   & 64.65\% & 62.10\%   & 67.60\% & 67.25\%   & 82.62\% & 80.11\% \\ 
    ~~Random   & Softmax   & 62.27\% & 56.64\%   & 67.58\% & 62.75\%   & 78.63\% & 77.17\% \\
    \multicolumn{8}{@{}l@{}}{\textbf{Regression}} \\ 
    ~~Places   & $L_2$     & 44.54\% & 45.86\%   & 46.84\% & 49.10\%   & 71.43\% & 69.70\% \\
    ~~Best     & $L_2$     & 55.54\% & 56.55\%   & 60.78\% & 62.16\%   & 76.65\% & 76.59\% \\
    ~~Places   & Huber     & 53.11\% & 53.85\%   & 57.79\% & 58.78\%   & 76.72\% & 76.72\% \\
    ~~Best     & Huber     & 62.86\% & 63.23\%   & 67.19\% & 67.27\%   & 81.19\% & 81.85\% \\
    \multicolumn{8}{@{}l@{}}{\textbf{Regression (regularized w/ classification)}} \\ 
    ~~Best     & $L_2$     & 57.29\% & 58.48\%   & 63.92\% & 64.41\%   & 79.24\% & 82.89\% \\
    ~~Best     & Huber     & 60.38\% & 60.51\%   & 67.18\% & 66.66\%   & 81.79\% & 82.55\% \\
    \multicolumn{8}{@{}l@{}}{\textbf{Other}} \\ 
    \multicolumn{2}{@{}l@{}}{~~Lezama et al.~\cite{lezama2014finding}} 
                           & \multicolumn{2}{c}{51.32\%} & \multicolumn{2}{c}{52.59\%} & \multicolumn{2}{c}{89.57\%} \\
    \multicolumn{2}{@{}l@{}}{~~Zhai et al.~\cite{zhai2016context}} 
                           & \multicolumn{2}{c}{57.33\%} & \multicolumn{2}{c}{58.24\%} & \multicolumn{2}{c}{90.80\%} \\
    \bottomrule
  \end{tabular}
  \label{tbl:classification}
\end{table}

When training a deep CNN, it is common practice to start optimization
from the weights of a previously trained
network~\cite{yosinski2014transferable} and ``fine-tune'' (updating
the weights of randomly initialized layers more). We apply this
strategy, in conjunction with our methods outlined in \secref{methods},
starting from a large number of pretrained CNNs. In all cases, we take
advantage of models made publicly available by the authors.  The
accuracy of each network on several datasets can be seen in in
\tblref{classification}, where the leftmost column represents which
network was used as initialization. We consider several different
initializations: a network trained for object recognition
(ImageNet~\cite{jia2014caffe}), a network trained for scene
categorization (Places~\cite{zhou2014places}), a network trained for
camera relocalization
(PoseNet-Street~\cite{kendall2015convolutional}), and a network
trained for salient object detection
(Salient~\cite{zhang2016unconstrained}).

As in \secref{comparison}, we compute horizon detection error and
report the area under the curve of the cumulative histogram of errors.
For classification, all networks have competitive performance on HLW,
but the choice of initialization is significant and we found the
$(\theta, \rho)$ parameterization to be superior. Our best network on HLW
achieves $69.97\%$ AUC using this parameterization and was initialized
using Places (we refer to this network as `Best' in the remainder of the
table). Overall performance is lower on the test imagery from the held
out models ({\em held}), compared to the full set ({\em all}). This result is
consistent with recent work on scene-specific camera
relocalization~\cite{kendall2015convolutional} demonstrating the
capability of a CNN to preserve pose information.

\begin{figure}
  \centering
  \includegraphics[height=.239\linewidth]{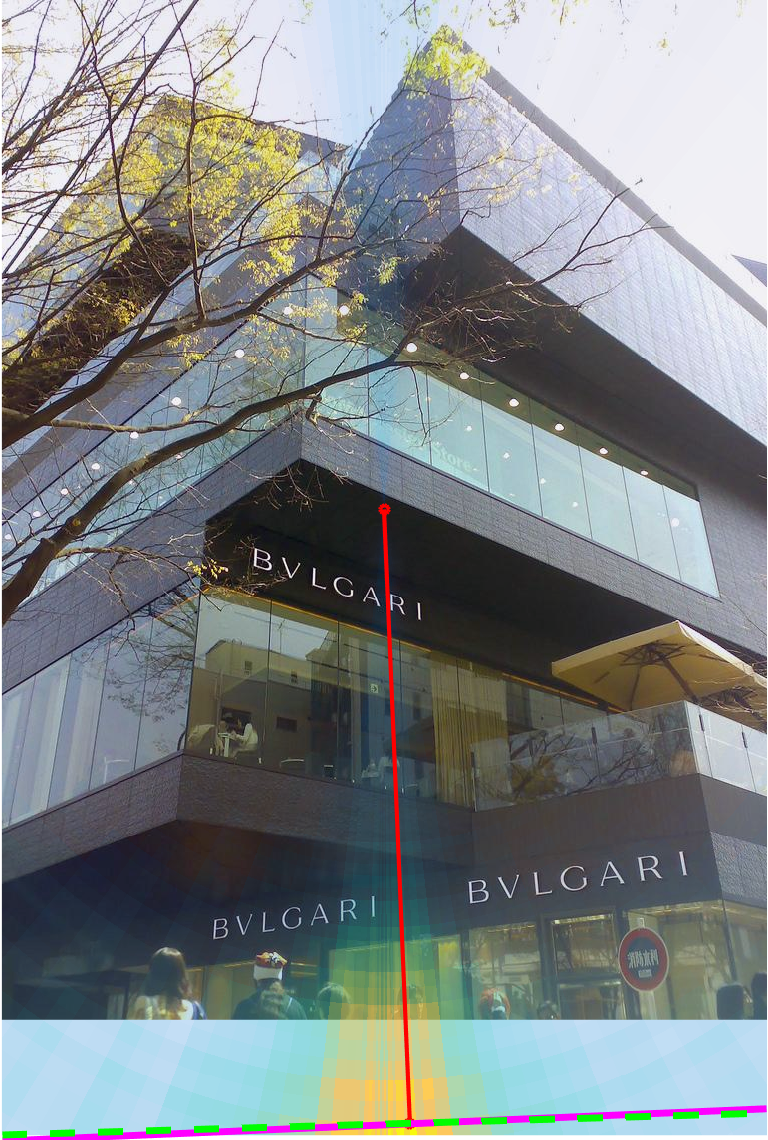}
  \includegraphics[height=.239\linewidth]{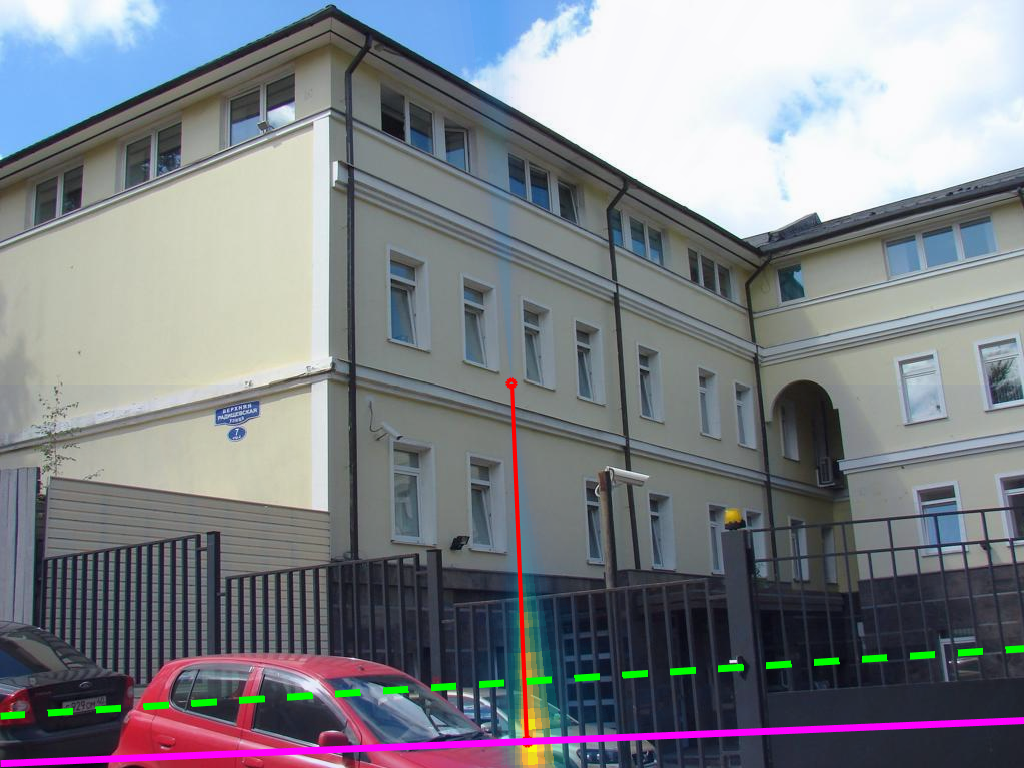}
  \includegraphics[height=.239\linewidth]{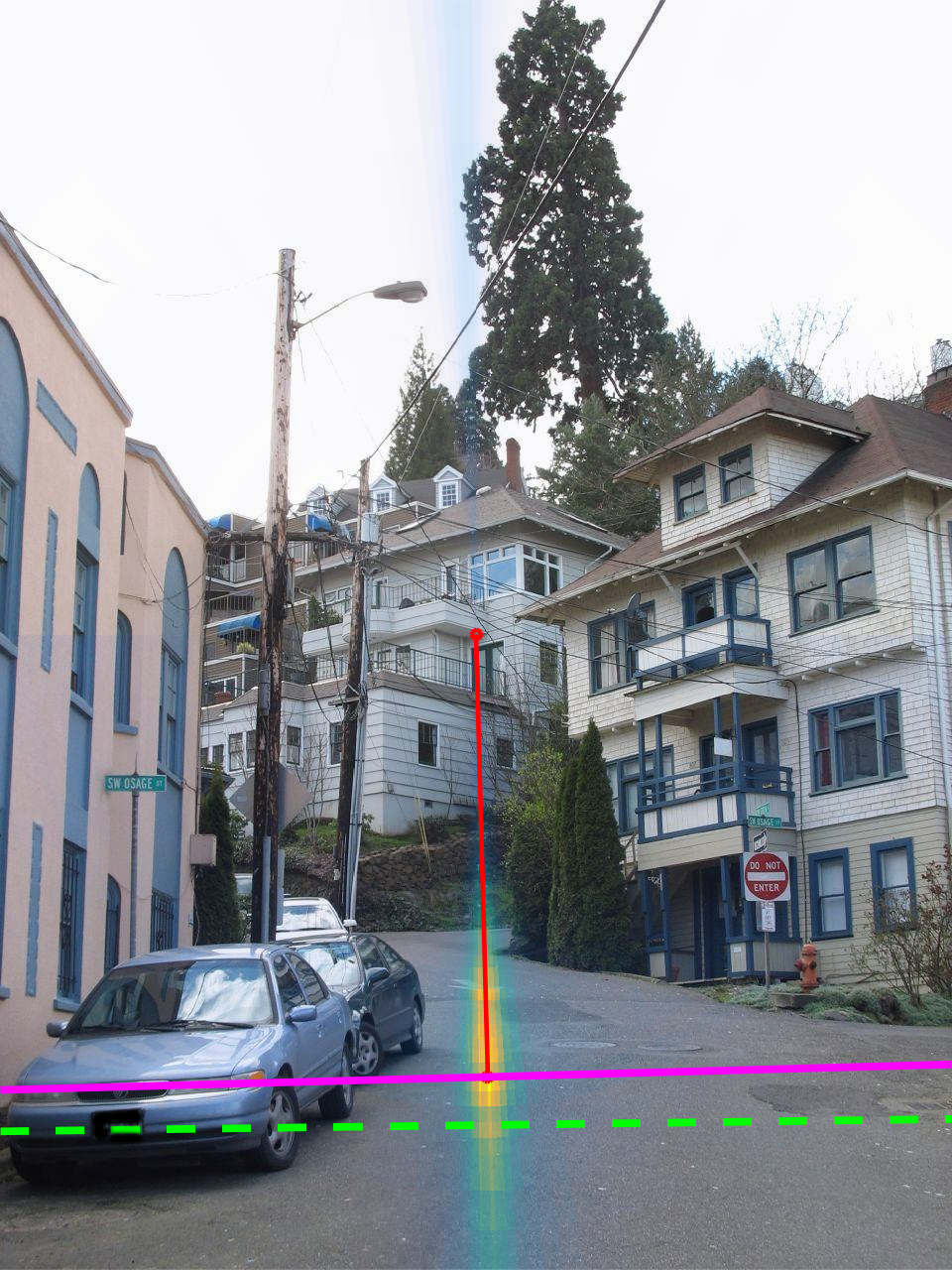}
  \includegraphics[height=.239\linewidth]{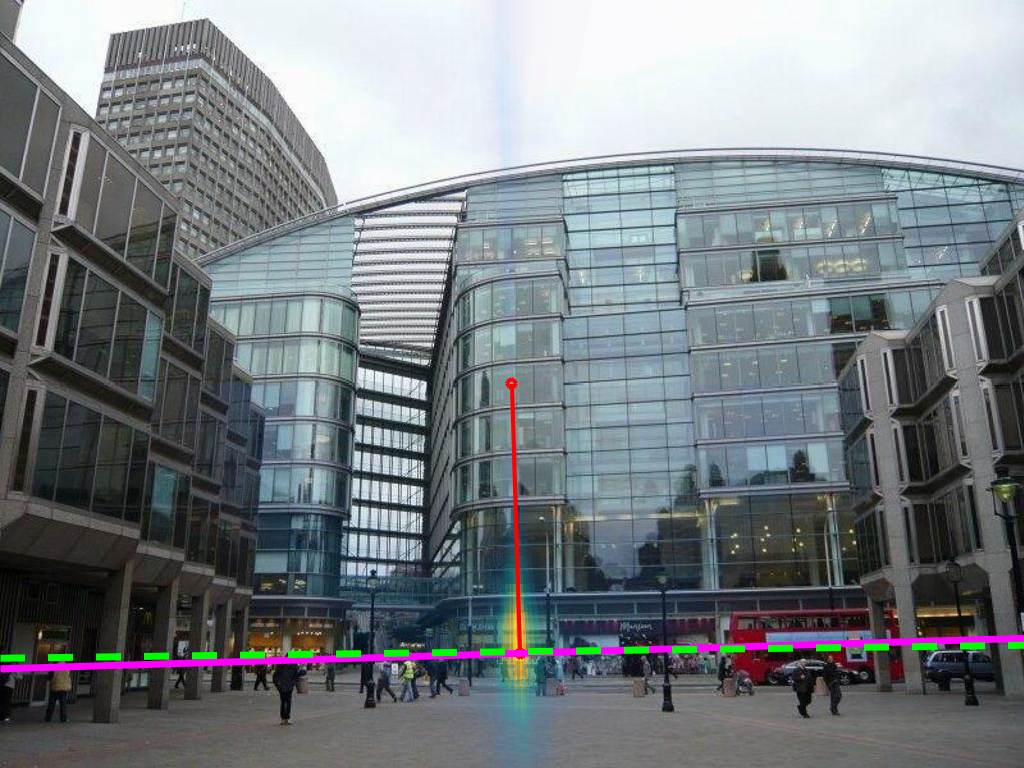}

  \caption{Example results showing the estimated distribution over
    horizon lines. For each image, the ground truth horizon line (dash
    green) and the predicted horizon line (magenta) are shown.  A
    false-color overlay (red = more likely, transparent = less likely)
    shows the estimated distribution over the point on the horizon
    line closest to the image center.}

  \label{fig:horizon_pdf}
\end{figure}

It proved more challenging to obtain good results for the regression
task. Fine-tuning performed much
worse than classification, for both loss functions, likely requiring
further manual parameter tweaking. Despite this, we found the $(l, r)$
parameterization to be superior, and the Huber loss to be
significantly better than the $L_2$ loss. Applying the strategies
outlined in \secref{regression}, namely initializing from the weights
of the best classification network and regularizing training with
softmax classifiers, significantly improves the performance of our
networks, making them competitive with classification. Qualitative
results from our approach are shown in \figref{horizon_pdf} for four
ECD images. 

Finally, using our best classification network we evaluate the
subwindow aggregation methods from \secref{merging}. The results are
shown in \tblref{additional}. In addition to a standard center crop,
we extract a $3 \times 3$ grid of crops (each $99\%$ of the minimum
dimension), chosen empirically using the HLW validation set. We saw no
benefit from using smaller crop sizes, as are commonly used for
semantic image classification. Both averaging in image space ({\em
average}) and optimizing across subwindows ({\em optimize})
significantly improve performance over a network evaluated on the
center crop alone. 

\begin{wraptable}{R}{.6\linewidth}
  \centering
  \caption{Evaluation of post-processing strategies.}
  \begin{tabular}{@{}lccc@{}}
    \toprule
    & HLW & ECD & YUD\\
    \hline
    Ours
    & 69.97\% & 83.96\% & 85.33\% \\
    Ours ({\em average}) 
    & \textbf{71.16\%} & 83.60\% & 86.41\% \\
    Ours ({\em optimize}) 
    & 70.66\% & 86.05\% & 86.11\% \\
    \midrule
    \cite{zhai2016context} (CNN = Orig.) 
      & 58.24\% & 90.80\% & 94.78\% \\
    \cite{zhai2016context} (CNN = Ours) 
    & 65.50\% & \textbf{91.29\%} & \textbf{95.46\%} \\
    \bottomrule
  \end{tabular}
  \label{tbl:additional}
\end{wraptable}

To highlight the ability of our networks, we update the recent
state-of-the-art method by Zhai et al.~\cite{zhai2016context}, which uses
a CNN to provide global context for vanishing point estimation, to use
our best classification network (using code provided by the authors).
With this change, we improve performance on HLW and advance the
state-of-the-art results on both the ECD and YUD datasets
(\tblref{additional}).  For ECD, our relative improvement in AUC is
$5.3\%$.  For YUD, our relative improvement is $13.0\%$, where Zhai et
al.~\cite{zhai2016context} previously reported a relative improvement
of $5.0\%$. Despite the limitations of these two benchmark datasets,
these are significant performance improvements.

\section{Conclusion}

We introduced {\em Horizon Lines in the Wild} (HLW), a new dataset for
single image horizon line estimation, to address the limitations of
existing horizon line detection datasets. HLW is several orders of
magnitude larger than any existing dataset for horizon line detection,
has a much wider variety of scenes and camera perspectives, and wasn't
constructed to highlight the value of any particular geometric cue.
Our hope is that it will continue to drive advances on this important
problem in the future. 

Using HLW, we investigated
methods for directly estimating the horizon line using convolutional
neural networks, including both classification and regression
formulations. Our methods are appealing because there is no need to
make explicit geometric assumptions on the contents of the
underlying scene, unlike virtually all existing methods, and we can
simultaneously take advantage of both geometric and semantic cues that
are present in the image. Despite this generality, the performance of
our methods is competitive, achieving state-of-the-art results on two
existing benchmark datasets designed for geometric methods, and
outperforming all existing methods on the challenging real-world
imagery contained in HLW. Our method is fast, works in natural
environments, and can provide a prior over horizon line location that
can be used as input to other methods.

\section*{Acknowledgements} 
We are grateful to Jan-Michael Frahm, Jared Heinly, Yunpeng Li,
Torsten Sattler, Noah Snavely, and Kyle Wilson for making SfM models
available to us.
This research was supported by the Intelligence Advanced
Research Projects Activity (IARPA) via Air Force Research Laboratory,
contract FA8650-12-C-7212.  The U.S. Government is authorized to
reproduce and distribute reprints for Governmental purposes
notwithstanding any copyright annotation thereon. Disclaimer: The
views and conclusions contained herein are those of the authors and
should not be interpreted as necessarily representing the official
policies or endorsements, either expressed or implied, of IARPA, AFRL,
or the U.S. Government.

\bibliography{biblio}

\end{document}